\pdfoutput=1

\documentclass[11pt]{article}

\usepackage{emnlp2021}

\usepackage{times}
\usepackage{latexsym}

\usepackage[T1]{fontenc}

\usepackage[utf8]{inputenc}

\usepackage{microtype}

\usepackage{amssymb,mathtools}
\usepackage{graphicx,paralist,multirow}
\usepackage{booktabs}

\title{Rethinking Why Intermediate-Task Fine-Tuning Works}

\author{\stepcounter{footnote}Ting-Yun Chang\thanks{Work was done when the first author was a research assistant at Academia Sinica, Taiwan.}\quad Chi-Jen Lu \\
    Institute of Information Science, Academia Sinica, Taiwan\\
  {\tt tingyun@usc.edu \quad cjlu@iis.sinica.edu.tw} 

  \\}

\begin{document}
\maketitle
\begin{abstract}
Supplementary Training on Intermediate Labeled-data Tasks (STILT) is a widely applied technique, which first fine-tunes the pretrained language models on an intermediate task before on the target task of interest.
While STILT is able to further improve the performance of pretrained language models, it is still unclear why and when it works. 
Previous research shows that those intermediate tasks involving complex inference, such as commonsense reasoning, work especially well for  RoBERTa-large.
In this paper, we discover that the improvement from an intermediate task could be orthogonal to it containing reasoning or other complex skills --- a simple real-fake discrimination task synthesized by GPT2 can benefit diverse target tasks. We conduct extensive experiments to study the impact of different factors on STILT. These findings suggest rethinking the role of intermediate fine-tuning in the STILT pipeline.
\end{abstract}

\section{Introduction}
\begin{figure}[t!]
  \centering
  \includegraphics[width=1\linewidth]{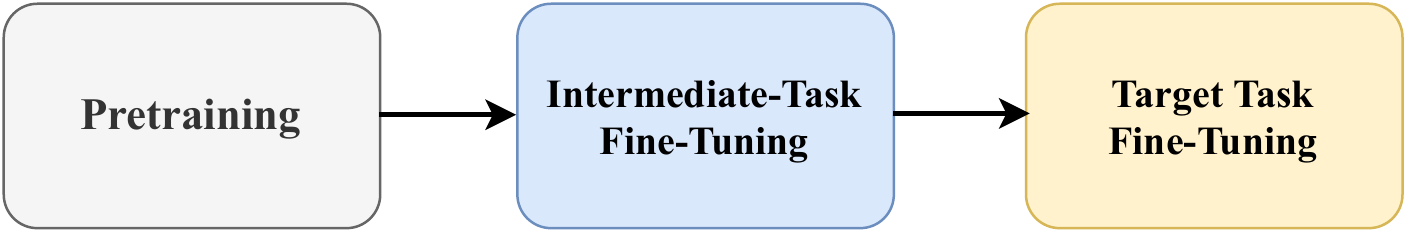}
  \caption{The pipeline of STILT.}
  \label{fig:STILT}
\end{figure}

Pretrained language models~\cite{peters2018deep, radford2018improving, devlin2019bert, liu2019roberta} have contributed to great progress in natural language understanding (NLU).
STILT~\cite{phang2018sentence, wang2019can, clark2019boolq, pruksachatkun2020intermediate, phang2020english, vu2020exploring} can further improve their performance on downstream NLU tasks by redesigning the training pipeline, introducing an intermediate-task fine-tuning phase before fine-tuning the pretrained models on the target task of interest (Figure~\ref{fig:STILT}). 
Nevertheless, this approach is not necessarily beneficial, and its effectiveness depends highly on the intermediate task applied. 

To study \emph{when} and \emph{why} STILT works, \citet{pruksachatkun2020intermediate} conduct large-scale experiments based on RoBERTa-large~\cite{liu2019roberta} with different intermediate-target task pairs.
They focus on studying what kind of intermediate tasks are helpful overall and which linguistic skills a model learns from the intermediate phase. They show the difficulty to have a generally useful intermediate task and conclude that those containing \emph{complex reasoning and inference}, such as CosmosQA~\cite{huang2019cosmos} and HellaSwag~\cite{zellers2019hellaswag}, tend to enhance various target tasks. However, this ignores the fact that HellaSwag is a synthetic dataset, and RoBERTa tends to capture the data artifacts when fine-tuned on HellaSwag~\cite{tamborrino2020pre}.

In this paper, we demonstrate that intermediate tasks' enhancement could be irrelevant to providing complex reasoning or special linguistic skills --- a simple real-fake discrimination task synthesized by GPT2~\cite{radford2019language} can benefit diverse target tasks, including those commonsense reasoning tasks.
These observations suggest rethinking the role of the intermediate-finetuning phase in the pipeline. Our main contributions are as follows.\footnote{Our source code is available at \url{https://github.com/terarachang/Rethinking_STILT.git}}
\begin{itemize}
\item We discover that a widely beneficial intermediate task is not required to provide specific linguistic or reasoning skills.
\item We highlight STILT's enhancement on fine-tuning stability, providing more than 1000 experimental observations on RoBERTa-large.
\item We study different factors that may influence STILT's efficacy, suggesting rethinking why it works.
\end{itemize}

\begin{table*}[t!]
  \small
  \centering
  \begin{tabular}{lccccc}
    \toprule
    \bf Name & \bf Size & \bf Task &  \bf Input Format & \bf Genre/Source \\
    \midrule
    HellaSwag  & 40k & sentence continuation & multiple-choice & ActivityNet, WikiHow \\
    HellaSwag-p & 40k & real-fake discrimination & multiple-choice & ActivityNet, WikiHow \\
    $\text{Synthesis}_{GPT2}$  & 30k & real-fake continuation & multiple-choice & Wikipedia \\
    \midrule
    CoLA  & 8.5k & linguistic acceptability & 1 sent. & linguistics publications \\
    WiC  & 5.4k & word sense disambiguation & 1 word; 2 sents. & WordNet, VerbNet... \\
    RTE  & 2.4k & natural language inference & 2 sents. & Wikipedia, news \\
    MedNLI & 11k & natural language inference & 2 sents. & MIMIC-III clinical notes \\
    SocialIQA & 33k & commonsense QA & multiple-choice & crowdsourcing \\
    $\text{WinoGrande}_{XS, M, L}$  & 0.2, 2.5, 10k & commonsense coreference & multiple-choice & crowdsourcing \\
    \bottomrule
  \end{tabular}
  \caption{Overview of the tasks in our experiments. We include more descriptions in the appendix.} 
  \label{table:tasks}
\end{table*}

\section{A \emph{Good} Intermediate Task} We first define what a good intermediate task means. STILT is known for two benefits~\cite{phang2018sentence}: 1) improving target tasks' \emph{best} performance, and 2) stabilizing the fine-tuning process of the target tasks, notably reducing the degenerate fine-tuning runs~\cite{devlin2019bert,dodge2020fine,mosbach2020stability}.
While~\citet{pruksachatkun2020intermediate} only focus on the first property, we study both benefits by extensive hyperparameter trials.
Summarized in Table~\ref{table:tasks}, we experiment on diverse, commonly used natural language understanding tasks, from word sense disambiguation to commonsense reasoning. A \emph{good} intermediate task should provide both benefits to these tasks.

Note that the definition of~\emph{stability} could be controversial. Here, we follow previous work~\cite{phang2018sentence,mosbach2020stability,dodge2020fine} in this research line and refer to "improving stability" as "reducing the variance of the validation performance", which is strongly related to "reducing the occurrence of degenerate runs over multiple hyperparameters trials" as the variance in performance is often dominated by degenerate runs.

\section{Rethinking: Two Simple Baselines} 
HellaSwag~\cite{zellers2019hellaswag} is a commonsense reasoning multiple-choice task, which contains a premise narrating an event and four plausible next scenarios (options) in each data example (Figure~\ref{fig:hella}). All its negative options are generated by the machine given the premises; consequently, this dataset is known to contain superfluous artifacts~\cite{tamborrino2020pre}.

Despite the limitation,~\citet{pruksachatkun2020intermediate} show that HellaSwag is one of the most potent intermediate tasks for RoBERTa-large in their large-scale experiments\footnote{They study RoBERTa with 110 intermediate-target task combinations and show that in many cases, the intermediate tasks are not helpful, or even hurtful in some cases.}. They then attribute such wide improvement on target tasks to the complex commonsense reasoning it requires. On the contrary, we first ablate the common sense from HellaSwag, seeking to understand if simple intermediate tasks are enough to enhance the performance of various target tasks. 

We propose two baselines as intermediate tasks. The first one is to \emph{remove the premises from HellaSwag}, denoted as HellaSwag-p, so that each data example only contains four options without contexts. Therefore, the model does not require common sense and reasoning skills to predict the follow-up anymore. It only needs to identify which option is \emph{not} generated by the machine.

Secondly, we \emph{build a synthetic dataset that mimics the creation of HellaSwag}, denoted as $\text{Synthesis}_{GPT2}$. The main difference is that, unlike HellaSwag, our premises and the correct endings are not from particular sources containing commonsense. We use Wikipedia as the source corpus since it has already been seen by the model in the pretraining phase. Specifically, given a sentence from Wikipedia, we split it into two parts. The first half becomes the premise, and the last half becomes the positive choice. We then use pretrained GPT2 to generate three negative choices conditioned on the premise. The decoding strategy is nucleus (top-$p$) sampling~\cite{holtzman2019curious}, where $p=0.9$. 
More descriptions can be found in the supplementary materials.

Our goal is to use these two simple baselines to point out some underestimated factors when studying why STILT works. While previous work~\cite{pruksachatkun2020intermediate} attempts to relate the linguistic skills between the intermediate and target tasks, we suspect that the linguistic knowledge further provided by the intermediate task could be less contributive than previous belief, as plenty of research on probing pretrained language models has shown that diverse linguistic skills are already learned in the pretrained models' representations~\cite{peters2018deep,tenney2018what, tenney2019bert, talmor2019olmpics}. 
Please note that instead of challenging common sense and complex reasoning can be good properties for an intermediate task, the proposed two baselines are meant to raise the need of rethinking other important aspects of what a beneficial intermediate task offers.

\begin{table}[t!]
  \centering
  \begin{tabular}{lcc}
    \toprule
    \multirow{2}{*}{Intermediate} & \multicolumn{2}{c}{Target Accuracy (\%)}\\
    \cline{2-3}
     & \bf RTE & \bf WiC \\
    \midrule
     \bf None & 83.5 / \underline{85.6} & 70.5 / \underline{71.8} \\
     \bf HellaSwag & 88.3 / \underline{88.4} & 70.6 / \underline{73.7} \\
  \bottomrule
  \end{tabular}
  \caption{Comparing~\protect\citeauthor{pruksachatkun2020intermediate}/our performance on the same intermediate-target task pairs.}
  \label{table:cmp}
\end{table}

\section{Experiments}
\subsection{Setup} Following~\citet{pruksachatkun2020intermediate}, we study the powerful pretrained model RoBERTa-large in all experiments. 
For each intermediate task, we perform a hyperparameter sweep over the learning rate in $\{5e-6,\ 1e-5,\ 2e-5\}$, the effective batch size in $\{8,\ 16,\ 32\}$, the warmup ratio in $\{0,\ 0.2\}$, and the random seed in $\{12,\ 42\}$\footnote{These two are the seeds recommended by~\citet{dodge2020fine} and used by \texttt{Huggingface} in default, respectively.} on every target task.\footnote{We use \texttt{Huggingface transformers} toolkit.}
That is, for each intermediate-target task pair, we conduct $3\times3\times2\times2=36$ experiments to study STILT's stability.
We follow the preprocessing of previous work. 
Due to some nuances in the setup and implementation, we first compare with~\citet{pruksachatkun2020intermediate} on the overlapped experiments in Table~\ref{table:cmp}, showing that our results are consistent with theirs.

\subsection{Results}
\label{sec:exp}
\begin{figure*}[t!]
  \centering
  \includegraphics[width=1\linewidth]{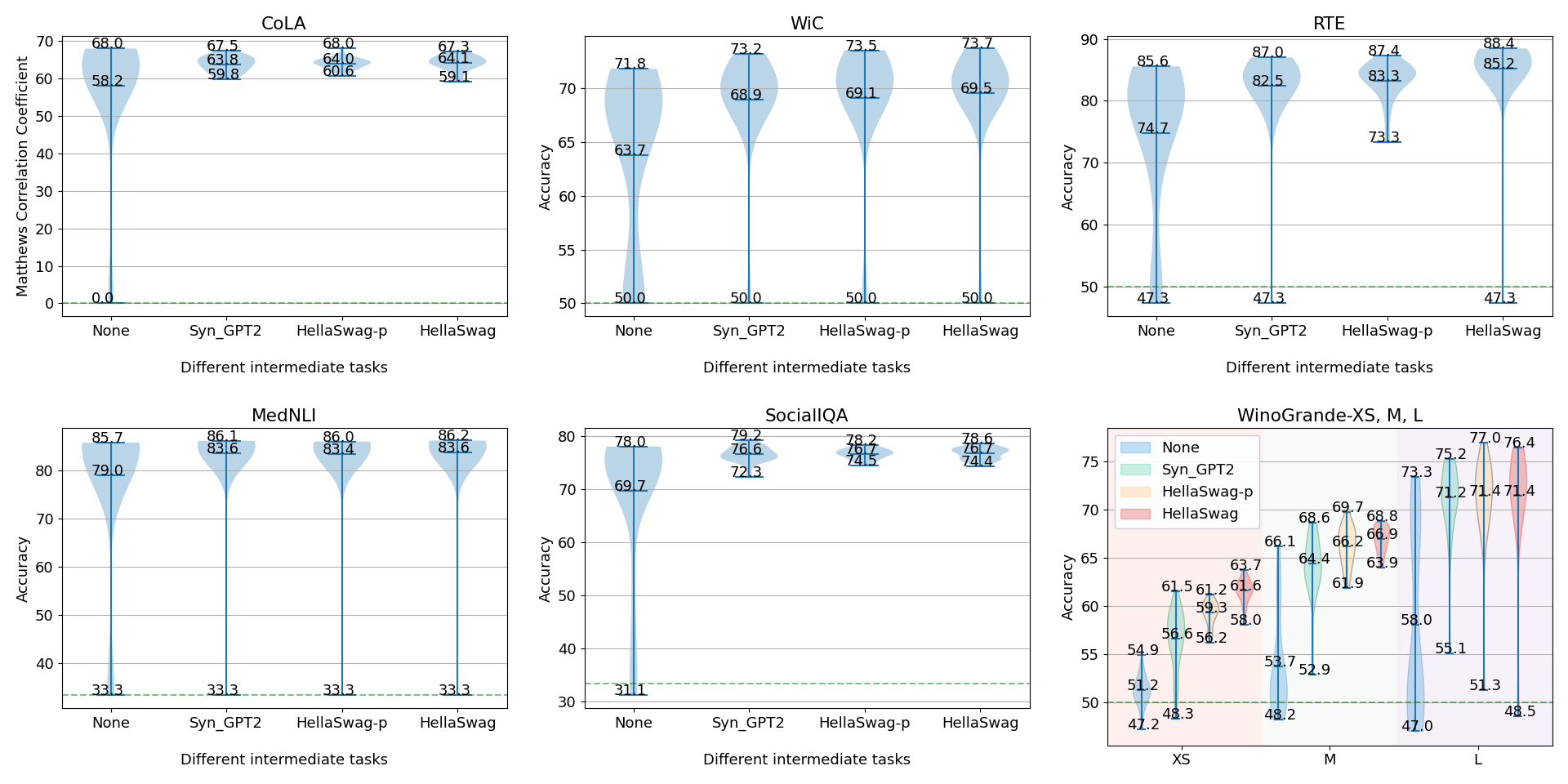}
  \caption{Results across different target tasks. Each \emph{violin} contains 36 hyperparameter trials of an intermediate-target pair, where the 3 annotated values correspond to the min, mean, and best performance within. The scores are $100\times$ for better visualization. The green dash line shows the random guessing performance on each task.}
  \label{fig:results}
\end{figure*}

Figure~\ref{fig:results} shows the experimental results on all the target tasks, where we use violinplot\footnote{\url{matplotlib.axes.Axes.violinplot}} to demonstrate results of all hyperparameters. Each subplot contains four methods (light blue \emph{violins}): 
\begin{itemize}
\item
\textbf{None}: not using any intermediate task, i.e., the standard, vanilla RoBERTa fine-tuning.
\item \textbf{HellaSwag}: using HellaSwag as the intermediate task.
\item \textbf{HellaSwag-p}: using the first proposed baseline, which ablates HellaSwag's premises.
\item \textbf{Syn\_GPT2}: using the second proposed intermediate task, which is synthesized by GPT2. 
\end{itemize}
We observe that \textbf{HellaSwag} does have generally\footnote{Note that~\emph{generally} does not mean~\emph{universally} works well.} positive effects compared with \textbf{None}, including enhancing the best performance and significantly reducing the degenerate runs on the various target tasks. 
To study what RoBERTa learns after fine-tuning on HellaSwag, we first test if it learns to select the endings according to the premises by removing all the premises in HellaSwag's dev set. The moderate drop in performance, from $84.8\%$ to $65.0\%$, where random guessing is only $25\%$, suggests that to some extent, it uses unwanted features in the machine-generated endings to make predictions.
Also, its zero-shot performance on the dev set of $\text{Synthesis}_{GPT2}$ is as high as $75.3\%$.
Thus, it is in doubt whether we can attribute \textbf{HellaSwag}'s improvement over \textbf{None} to offering RoBERTa commonsense reasoning skills.

\begin{table}[t!]
  \label{tab}
  \centering
  \begin{tabular}{lrrr}
    \toprule
    \multirow{2}{*}{Interm.} & \multicolumn{3}{c}{$\Delta$ Mean/Best Accuracy (\%)}\\
    \cline{2-4} & \multicolumn{1}{c}{\bf MedNLI} & \multicolumn{1}{c}{\bf WiC} & \multicolumn{1}{c}{\bf $\text{WinoG}_M$} \\
    \midrule
     \bf CoLA & +1.3 / +0.2 & +3.9 / \textcolor{red}{-0.3} & \textcolor{red}{-1.6} / \textcolor{red}{-3.8} \\  
     \bf Hella-sh & +1.4 / \textpm 0.0 & +2.8 / \textcolor{red}{-0.9}  & \textcolor{red}{-1.7} / \textcolor{red}{-3.5} \\
  \bottomrule
  \end{tabular}
  \caption{The effect of using other true-false tasks, CoLA and Hella-sh, as the intermediate tasks.}
  \label{table:more}
\end{table}

Meanwhile, the proposed baselines show a competitively positive effect across all target tasks, including those in specific domains, such as SocialIQA (commonsense) and MedNLI (medical).
As our simple baselines do not contain knowledge in these fields or other linguistic skills\footnote{One could argue that the two baselines still include some \emph{reasoning} skills, which may lead to a long debate, depending on the definition of \emph{reasoning}. Another debatable issue is that whether our baselines are really~\emph{simple}.} besides real-fake discrimination, the overall improvement requires a careful rethinking of why STILT works. 

\section{Analysis}
In this section, we further study different factors that may influence the effectiveness of STILT.
\subsection{Intermediate Tasks}
While we demonstrate the efficacy of the two simple baselines in the previous section, here we investigate if other true-false intermediate tasks also work, including 1) CoLA, a task of grammaticality, and 2) Hella-sh, a dummy task created by shuffling the words in the fake endings of HellaSwag. 
Table~\ref{table:more} shows that they both contribute \emph{negatively} (red-colored) in many cases. 
We suppose that a true-false intermediate task works widely when it provides RoBERTa general, high-level overlaps with target tasks. For example, focusing on summarizing the semantic-level information to the [CLS] token so that the classifier atop can make decisions easier since RoBERTa's pretraining only applies mask language modeling~\cite{liu2019roberta}.
On the contrary, leaning toward learning specific rules or skills such as linguistic acceptability (CoLA) cannot benefit diverse target tasks.
In this paper, we raise the need for rethinking by showing the different efficacy of some related intermediate tasks and leave it for future work to provide a more convincing explanation on why or why not they work.

\subsection{Target Training Size}
\label{sec:size}
Previous work~\cite{phang2018sentence, pruksachatkun2020intermediate, vu2020exploring} has found that STILT works especially well on limited labeled target tasks. Here, we study the impact of target-task size on $\text{WinoGrande}_{XS, M, L}$~\cite{sakaguchi2019winogrande}, since the dataset contains different training sizes.
Figure~\ref{fig:results} shows that RoBERTa can barely learn from the 160 training data of $\text{WinoGrande}_{XS}$ with vanilla fine-tuning. We observed that the training loss was about constant during the entire fine-tuning phase. At this point, introducing the intermediate tasks notably enhances the model's stability and its best performance. 
However, when we increase the training size, the improvements on the best performance dwindle. 
On the other hand, the average-performance improvements remain significant, mainly because RoBERTa still suffers from a few degenerate runs. 
This section shows that the target-task size has a strong influence on STILT's effectiveness, especially when the pretrained model struggles to learn from the sparse training signals, where STILT can help converge better.

\begin{figure}[t!]
  \centering
  \includegraphics[width=1\linewidth]{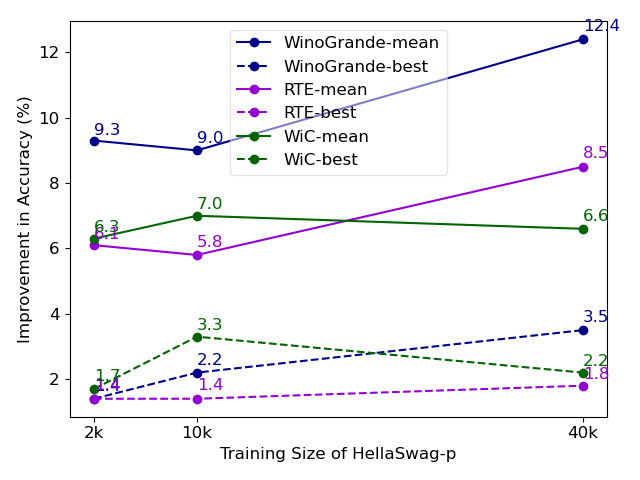}
  \caption{The influence of the intermediate-task training size (2k, 10k, 40k). We run a hyperparameter sweep for each size and report the best (solid lines) and mean (dashed lines) improvements over \textbf{None}.}
  \label{fig:inter_sz}
\end{figure}

\subsection{Intermediate Training Size}
Finally, we study the influence of the intermediate-task training size on three target tasks: $\text{WinoGrande}_M$, RTE, and WiC. Figure~\ref{fig:inter_sz} shows that fine-tuning on a few data (2000) of the intermediate task, \textbf{HellaSwag-p}, already leads to noticeable improvements on these target tasks. 
We suggest rethinking what the intermediate task provides under such a few-resource circumstance.
We believe that instead of providing RoBERTa more linguistic knowledge related to the target tasks, the intermediate task offers some high-level guidance to bridge the gaps between the pretraining and fine-tuning phases.

\section{Conclusion}
We discover that a generally beneficial intermediate task to RoBERTa can be as simple as a synthetic real-fake discrimination task, and provide observations on different factors that influence STILT's best and mean effectiveness. 
Therefore, we suggest rethinking why intermediate-task fine-tuning works, particularly under low-resource settings.

\section{Acknowledgments} We would like to thank Ting-Rui Chiang, Ta-Chung Chi, and Jian-Jia Su for in-depth discussions. We are thankful to the anonymous reviewers for their insightful comments on the paper.

\bibliography{anthology,custom}
\bibliographystyle{acl_natbib}

\appendix

\clearpage
\section{Datasets Details}
\subsection{Intermediate Tasks}
In this section, we include more details about the HellaSwag dataset and how we construct our second baseline, $\text{Synthesis}_{GPT2}$.

\begin{figure}[t!]
  \centering
  \includegraphics[width=1.0\linewidth]{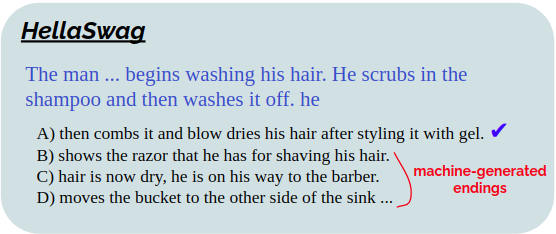}
  \caption{An example in HellaSwag dataset.}
  \label{fig:hella}
\end{figure}

\textbf{HellaSwag} is a commonsense reasoning task that tests a model’s ability to choose the most plausible continuation of an event. The premises and the correct options are derived from ActivityNet Captions~\cite{krishna2017dense} and WikiHow to include commonsense knowledge, while its negative options are GPT-generated. Adversarial Filtering~\cite{zellers2018swag,zellers2019hellaswag} are applied against BERT to create more challenging options. Figure~\ref{fig:hella} illustrates an example in HellaSwag.

In our $\text{Synthesis}_{GPT2}$ baseline, we mimicked HellaSwag's creation process to build a sentence continuation task \emph{without} commonsense knowledge for the ablation study. 
We chose Wikipedia as the source of the premises and correct answers, while the negative options are generated by GPT2-medium. 
We did not apply Adversarial Filtering.

Please note that we only run hyperparameter sweeps on the target tasks, not on the intermediate tasks, as we believe that a handy intermediate task should not require the resource-consuming hyperparameter search. 
For intermediate tasks, we simply use the central hyperparameters in the span (learn rate=1e-5, batch size=16, random seed=42).

\subsection{Target Tasks}
Here, we make a brief introduction about the target tasks we evaluate on.

\textbf{CoLA} The Corpus of Linguistic Acceptability~\cite{warstadt2019neural} is a binary classification task containing sentences labeled as either grammatical or ungrammatical. Performance on CoLA is reported in Matthew’s correlation coefficient (MCC). CoLA is a task in GLUE benchmark.

\textbf{RTE} Recognizing Textual Entailment~\cite{dagan2005pascal} is a textual entailment task. 
We use the binary sentence
classification version of the task. 
Each example contains a premise and a hypothesis sentence. 
Performance on RTE is reported in accuracy. 
RTE is a task in both GLUE and SuperGLUE benchmarks.

\textbf{WiC} Word-in-Context~\cite{pilehvar2019wic} is a binary classification word sense disambiguation task. 
Each example consists of two sentences and a polysemous word that appears in both sentences, asking whether the word has the same sense in both. 
Performance on WiC is reported in accuracy. 
WiC is a task in SuperGLUE benchmark.

\textbf{WinoGrande} An Adversarial Winograd Schema Challenge at Scale~\cite{sakaguchi2019winogrande} is a commonsense coreference resolution task, which improves the scale and the hardness of WSC~\cite{levesque2012winograd}. 
Each example contains one sentence with a blank and two options to be filled in. 
We follow~\citet{sakaguchi2019winogrande} when preprocessing its input for RoBERTa. 
For example, an instance is formatted as \emph{" [CLS] The trophy doesn't fit into the brown suitcase because the [SEP] \_ is too large. [SEP]"}, where the blank is filled with either option1 or option2. 
This dataset includes different training scales, where we use the XS, M, and L versions in this paper.
Performance on WinoGrande is reported in accuracy. 

\textbf{SocialIQA}~\cite{sap2019socialiqa} is a multiple-choice commonsense question-answering dataset.
Each example consists of a context, a question, and three options. 
The task is about commonsense reasoning that requires emotional and social intelligence in everyday situations. 
Performance on SocialIQA is reported in accuracy.

\textbf{MedNLI}~\cite{romanov2018lessons} is a natural language inference dataset for the clinical domain, which is annotated by doctors and grounded in the medical history of patients. 
The premise sentences are from MIMIC-III~\cite{johnson2016mimic}. 
The label classes are \emph{entailment}, \emph{contradiction}, and \emph{neutral}. 
Performance on MedNLI is reported in accuracy. 
MedNLI is a task in PhysioNet~\cite{PhysioNet}.

\begin{table}[t!]
  \centering
  \begin{tabular}{lcc}
    \toprule
    \multirow{2}{*}{Intermediate} & \multicolumn{2}{c}{Zero-shot Acc. (\%)}\\
    \cline{2-3}
     & \bf SocialIQA & \bf WinoGrande \\
    \midrule
     \bf None & 35.0 & 52.3 \\
     \cline{1-3}
     \bf Syn\_GPT2 & 43.2 & 54.3 \\
     \bf HellaSwag-p & 44.0 & 56.4 \\
  \bottomrule
  \end{tabular}
  \caption{Zero-shot performance on two target tasks: SocialIQA and WinoGrande.}
  \label{table:zero}
\end{table}

\section{Future Work} Our work raises the need for rethinking why intermediate fine-tuning works. 
We found that in some target tasks, STILT's efficacy seems to be correlated with the phenomenon of degenerate fine-tuning runs~\cite{devlin2019bert,dodge2020fine,mosbach2020stability}. 
Thus, more research in degenerate runs may help us better understand how STILT works.

Unfortunately, we are unable to provide a convincing explanation on why and how our simple baselines work across target tasks in various domains.
We suspect that after the large-scale pretraining, RoBERTa-large has already learned a certain amount of knowledge required in the downstream target tasks and that our proposed intermediate tasks work well as they help RoBERTa bridge the gaps between the pretraining and fine-tuning phases. 
For example, they probably help summarize the semantic-level information to the [CLS] token so that the classifier atop can make decisions easier since RoBERTa's pretraining only applies mask language modeling. 
However, this is just an unverified hypothesis.

We conduct an experiment related to our hypothesis. 
We evaluate our two baselines on the dev sets of SocialIQA and WinoGrande without fine-tuning on their training sets. 
We can apply such a zero-shot setting as our baselines share the same model architecture, \texttt{RobertaForMultipleChoice}\footnote{\url{https://huggingface.co/transformers/model_doc/roberta.html\#robertaformultiplechoice}}, with WinoGrande and SocialIQA.\footnote{Similarly, we cannot conduct such experiments on other target tasks as they do not share the same architecture.}
The results in Table~\ref{table:zero} show that our simple intermediate fine-tuning methods, Syn\_GPT2 and HellaSwag-p, have better performance over pretrained RoBERTa (None), although they can hardly provide RoBERTa with the commonsense knowledge required in SocialIQA and WinoGrande. Where does the improvement come from? Could it give credence to our hypothesis that bridging the gap between pretraining and downstream tasks?

Besides, we acknowledge that one could argue that the two baselines still include some \emph{reasoning} skills, depending on the definition of \emph{reasoning}.
We believe that after the research community formulates clear notions and definitions on \emph{reasoning} and \emph{common sense}, we can have a better understanding of STILT.
Similarly, another debatable issue is that whether our baselines are really~\emph{simple}. In this paper, we only meant to show that proposed strong baselines are not heavily human curated and unintuitively work well.

We leave it for future work to better understand when, why, and how STILT can help what target task. We believe that recent work in pretrained language models' transferability~\cite{vu2020exploring, tamkin2020investigating, zhang2020revisiting, chung2020rethinking} can provide some insights into these questions. 

\section{Implementation Details} 
All our models are based on \texttt{HuggingFace's transformers} \texttt{Pytorch} toolkit. We use \texttt{RobertaForSequenceClassification} class for RTE, CoLA, MedNLI, and WiC; and use \texttt{RobertaForMultipleChoice} for WinoGrande, SocialIQA, HellaSwag, and our two baselines. The RoBERTa-large model contains 24-layer, 1024-hidden, and 16-heads, with $\sim350$M parameters totally.

\end{document}